\documentclass[10pt,twocolumn,letterpaper]{article}

\usepackage{wacv}
\usepackage{times}
\usepackage{graphicx}
\usepackage{amsmath}
\usepackage{amssymb}
\usepackage{balance}
\usepackage{url}
\usepackage{mathtools}
\usepackage{xspace}
\usepackage{multirow}
\usepackage{booktabs}
\usepackage{algorithmic}
\usepackage{algorithm}

\graphicspath{{./}{./figures/}}

\wacvfinalcopy 

\def\wacvPaperID{3} 

\def\Vec#1{{\boldsymbol{#1}}}
\def\Mat#1{{\boldsymbol{#1}}}

\newcommand{\tr}{\mbox{tr}}

\ifwacvfinal\fi
\begin{document}

\title
  {
 Object Tracking via Non-Euclidean Geometry: A Grassmann Approach
  }

\author
  {
  {\it Sareh Shirazi, Mehrtash T. Harandi, Brian C. Lovell, Conrad Sanderson}\\
  ~\\
  NICTA, GPO Box 2434, Brisbane, QLD 4001, Australia\\
  University of Queensland, School of ITEE, QLD 4072, Australia\\
  Queensland University of Technology, Brisbane, QLD 4000, Australia
  }

\maketitle
\thispagestyle{empty}

\begin{abstract}
\vspace{-1.5ex}

\noindent
A robust visual tracking system requires an object appearance model
that is able to handle occlusion, pose, and illumination variations in the video stream.
This can be difficult to accomplish when the model is trained using only a single image.
In this paper, we first propose a tracking approach based on affine subspaces (constructed from several images)
which are able to accommodate the abovementioned variations.
We use affine subspaces not only to represent the object, but also the candidate areas that the object may occupy.
We furthermore propose a novel approach to measure affine subspace-to-subspace distance via the use of non-Euclidean geometry of Grassmann manifolds.
The tracking problem is then considered as an inference task in a Markov Chain Monte Carlo framework via particle filtering.
Quantitative evaluation on challenging video sequences indicates that the proposed approach
obtains considerably better performance than several recent state-of-the-art methods
such as Tracking-Learning-Detection and MILtrack.

\end{abstract}

\vspace{-2ex}
\section{Introduction}
\label{sec:introduction}

Visual tracking is a fundamental task in many computer vision applications
including event analysis, visual surveillance, human behaviour analysis, and video retrieval~\cite{li2012incremental}.
It is a challenging problem, mainly because the appearance of tracked objects changes over time.
Designing an appearance model that is robust against intrinsic object variations
(\eg shape deformation and pose changes)
and extrinsic variations
(\eg camera motion, occlusion, illumination changes)
has attracted a large body of work~\cite{babenko2011,wang2011superpixel}.

Rather than relying on object models based on a single training image,
more robust models can be obtained through the use of several images,
as evidenced by the recent surge of interest in object recognition techniques
based on image-set matching.
Among the many approaches to image-set matching,
superior discrimination accuracy, as well as increased robustness to practical issues (such as pose and illumination variations),
can be achieved by modelling image-sets as linear subspaces~\cite{Harandi_2013_ICCV,harandi2011graph,Harandi_2013_PRL,Sanderson_AVSS_2012,Shirazi_2012_ICIP,turaga2011statistical}.

In spite of the above observations, we believe modelling via linear spaces is not completely adequate for object tracking.
We note that all linear subspaces of one specific order have a common origin.
As such, linear subspaces are theoretically robust against translation,
meaning a linear subspace extracted from a set of points does not change if the points are shifted equally.
While the resulting robustness against small shifts is attractive for object recognition purposes,
the task of tracking is to generally maintain precise locations of objects.

\begin{figure*}[!tb]

  \vspace{3.5ex}

  \centering
  \begin{minipage}{1.0\textwidth}
    \centering
    \begin{minipage}{0.32\textwidth}
      \centering
      \includegraphics[width=1\textwidth,keepaspectratio]{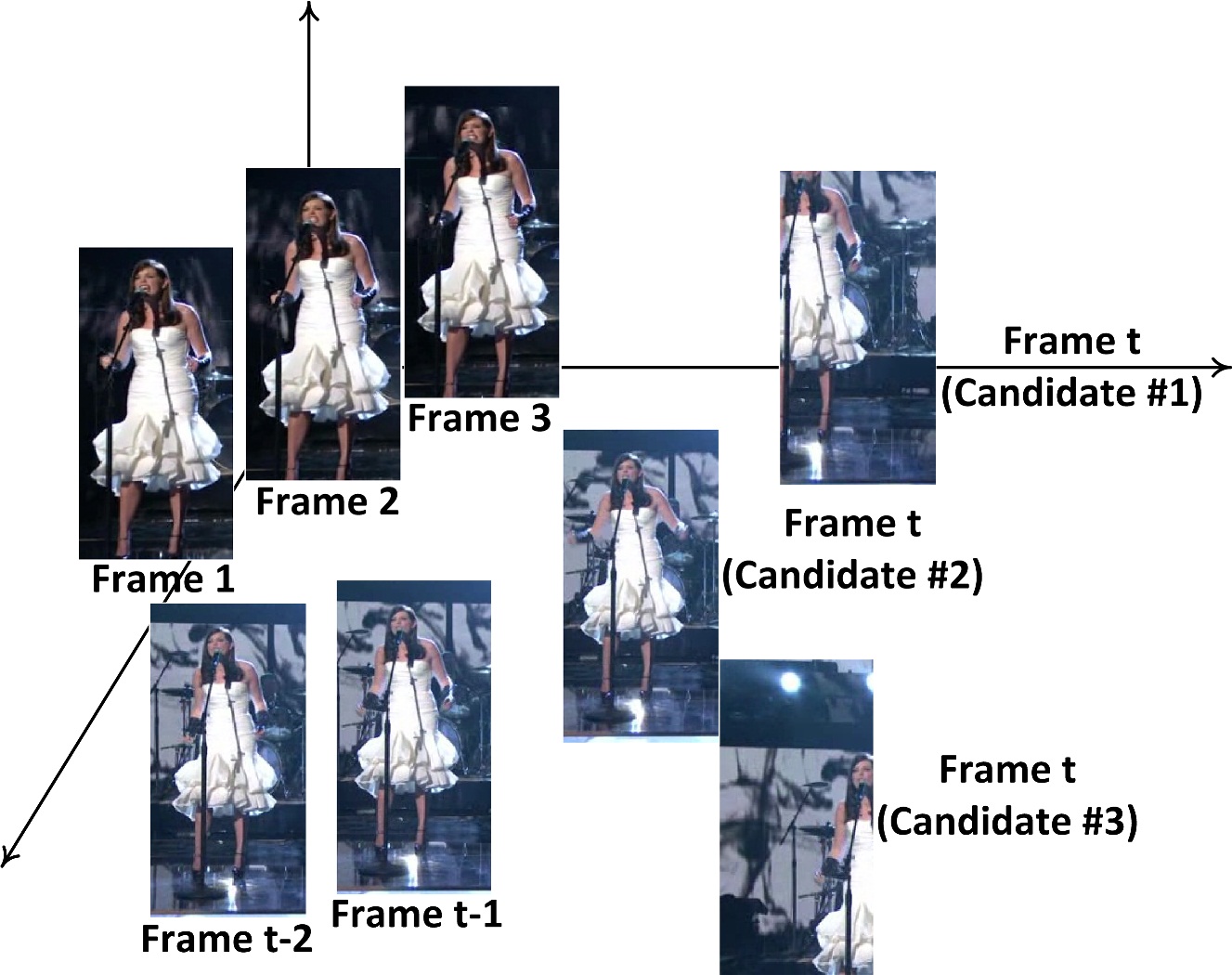}
    \end{minipage}
    %
    %
    \begin{minipage}{0.32\textwidth}
      \centering
      \includegraphics[width=1\textwidth,keepaspectratio]{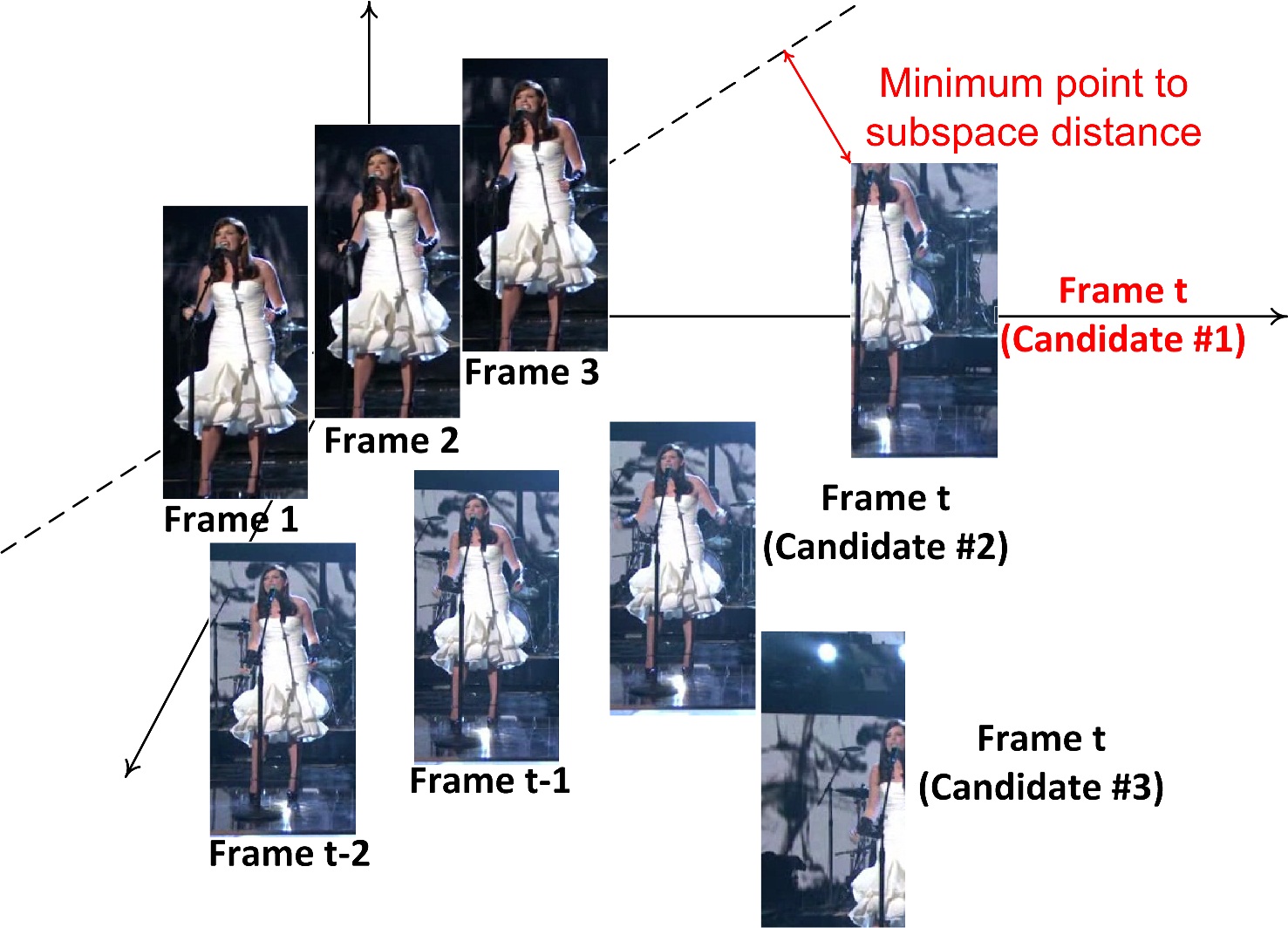}
    \end{minipage}
    ~
    \begin{minipage}{0.32\textwidth}
      \centering
      \includegraphics[width=1\textwidth,keepaspectratio]{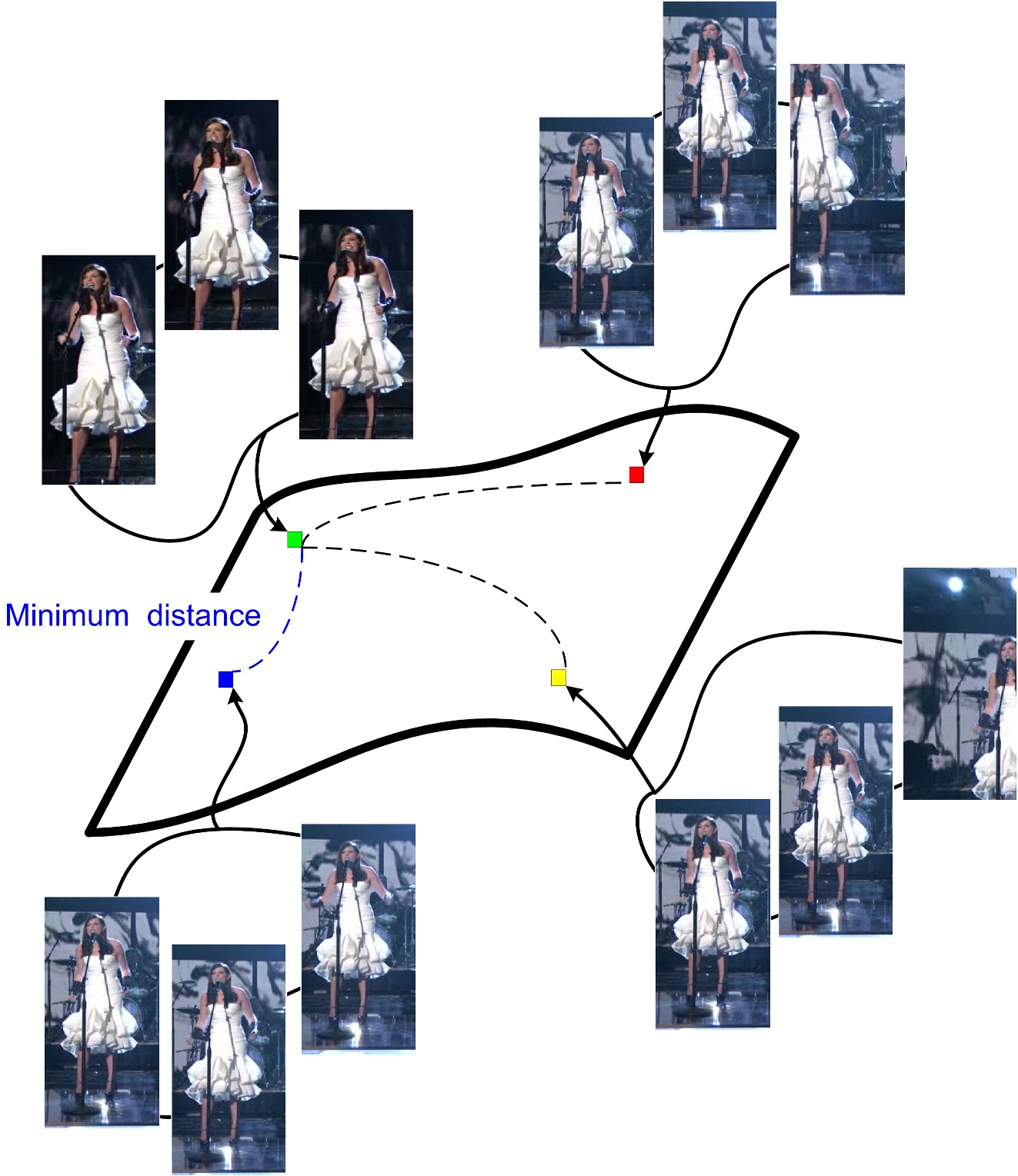}
    \end{minipage}
   %

    %
    \begin{minipage}{0.32\textwidth}
      \centering
      {\bf (a)}
    \end{minipage}
    %
    %
    \begin{minipage}{0.32\textwidth}
      \centering
      {\bf (b)}
    \end{minipage}
    \hfill
    \begin{minipage}{0.32\textwidth}
      \centering
      {\bf (c)}
    \end{minipage}
  \end{minipage}

  \vspace{1ex}

  \caption
    {
    Difference between point-to-subspace and subspace-to-subspace distance measurement approaches.
    {\bf (a)}~Three groups of images, with each image represented as a point in space;
              the first group (top-left) contains three consecutive object images (frames 1, 2 and 3) used for generating the object model;
              the second group (bottom-left) contains tracked object images from frames $t-2$ and $t-1$;
              the third group (right) contains three candidate object regions from frame $t$.
    {\bf (b)}~Subspace generated based on object images from frames 1, 2 and 3, represented as a dashed line;
              the minimum point-to-subspace distance can result in selecting the {wrong} candidate region (\ie wrong location).
    {\bf (c)}~Generated subspaces, represented as points on a Grassmann manifold;
              the top-left subspace represents the object model;
              each of the remaining subspaces was generated by using tracked object images from frames $t-2$ and $t-1$,
              with the addition of a unique candidate region from frame $t$;
              using subspace-to-subspace distance is more likely to result in selecting the correct candidate region.
    }
  \label{fig:pointsubspace}
  \vspace{0.5ex}
  \hrule

  \vspace{-2ex}

\end{figure*}

To account for the above problem, in this paper we first propose to model objects,
as well as candidate areas that the objects may occupy,
through the use of generalised linear subspaces, \ie affine subspaces, where the origin of subspaces can be varied.
As a result, the tracking problem can be seen as finding the most similar affine subspace
in a given frame to the object's affine subspace.
We furthermore propose a novel approach to measure distances between affine subspaces,
via the use of non-Euclidean geometry of Grassmann manifolds,
in combination with Mahalanobis distance between the origins of the subspaces.
See Fig.~\ref{fig:pointsubspace} for a conceptual illustration of our proposed distance measure.

To the best of our knowledge,
this is the first time that appearance is modelled by affine subspaces for object tracking.
The proposed approach is somewhat related to adaptive subspace tracking~\cite{ho2004visual,ross2008incremental}.
Ho \etal~\cite{ho2004visual} represent an object as a point in a linear subspace, which is constantly updated.
As the subspace was computed using only recent tracking results,
the tracker may drift if large appearance changes occur.
In addition, the location of the tracked object is inferred via measuring point-to-subspace distance,
which is in contrast to the proposed method, where a more robust subspace-to-subspace distance is used.

Ross \etal~\cite{ross2008incremental} improved tracking robustness against large appearance changes
by modelling objects in a low-dimensional subspace,
updated incrementally using all preceding frames.
Their method also involves a point-to-subspace distance measurement to localise the object.

The proposed method should not be confused with subspace learning on Grassmann manifolds proposed by Wang \etal~\cite{wang2008online}.
More specifically, in~\cite{wang2008online} an online subspace learning scheme using Grassmann manifold geometry
is devised to learn/update the subspace of object appearances.
In contrast to the proposed method, they also consider the point-to-subspace distance to localise objects.

\section{Proposed Affine Subspace Tracker (AST)}
\label{sec:model_v3}

\begin{figure}[!t]
  \centering
  \includegraphics[width=0.95\columnwidth,height=0.58\columnwidth]{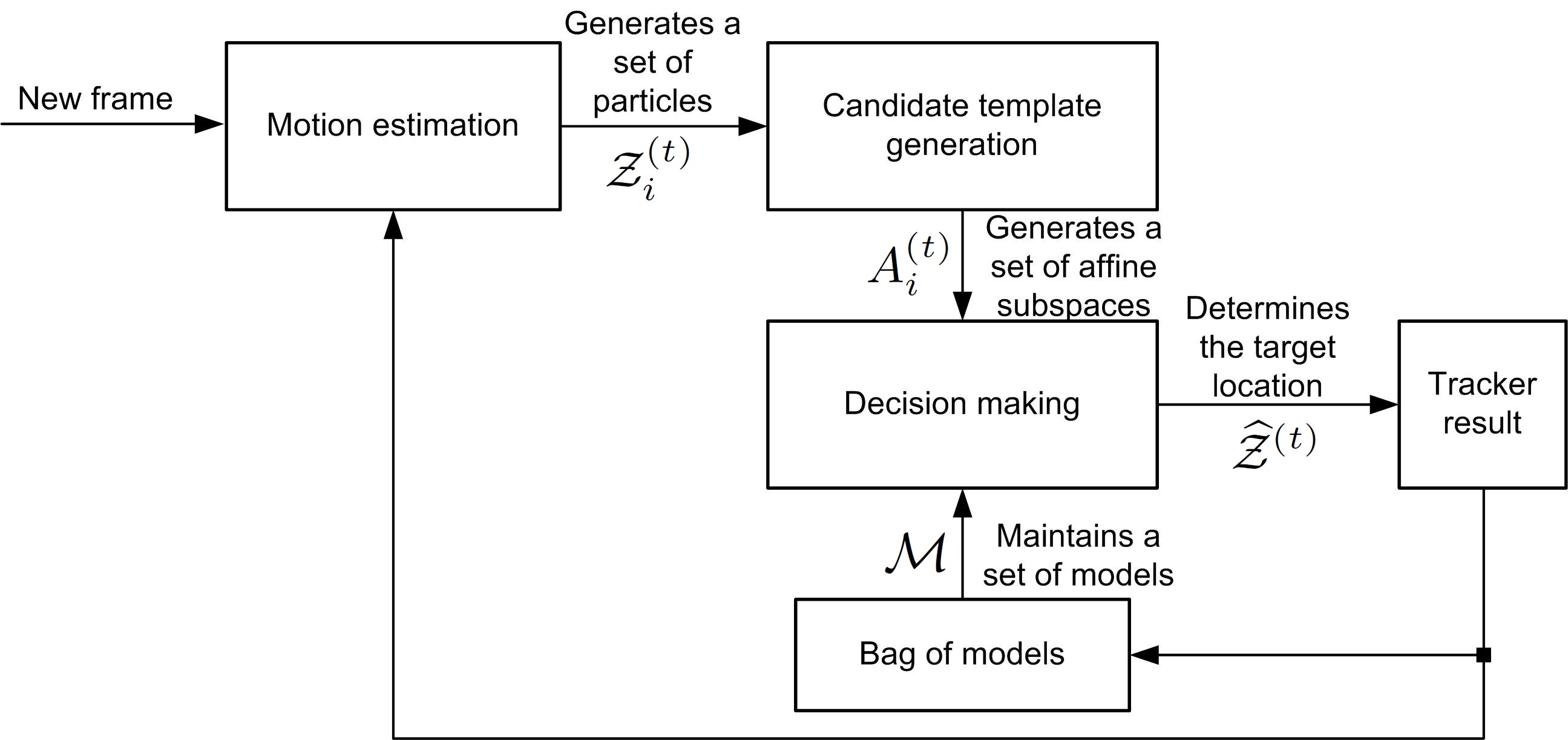}
  \caption
    {
    Block diagram for the proposed Affine Subspace Tracker (AST).
    }
  \label{fig:Bkd}
  \vspace{-1ex}
\end{figure}

The proposed Affine Subspace Tracker (AST) is comprised of four components, overviewed below.
A block diagram of the proposed tracker is shown in Fig.~\ref{fig:Bkd}.

\begin{enumerate}

\item
{\bf Motion Estimation}.
This component takes into account the history of object motion in previous frames and creates a set of candidates as to where the object might be found in the new frame.
To this end, it parameterises the motion of the object between consecutive frames as a distribution via particle filter framework~\cite{maskell2001tutorial}.
Particle filters
are sequential Monte Carlo methods and use a set of points to represent the distribution.
As a result, instead of scanning the whole of the new frame to find the object, only highly probable locations will be examined.

\item
{\bf Candidate Subspaces}.
This module encodes the appearance of a candidate (associated to a particle filter) by an affine subspace $A_{i}^{(t)}$.
This is achieved by taking into account the history of tracked images and learning the
origin $\mu_{i}^{(t)}$ and basis $U_{i}^{(t)}$ of $A_{i}^{(t)}$ for each particle.

\item
{\bf Decision Making}.
This module measures the likelihood of each candidate subspace $A_{i}^{(t)}$ to the stored object models in the bag $\mathcal{M}$.
Since object models are encoded by affine subspaces as well, this module determines the similarity between affine subspaces. The most similar candidate subspace to the bag $\mathcal{M}$ is selected as the result of tracking.

\item
{\bf Bag of Models}.
This module keeps a history of previously seen objects in a bag.
This is primarily driven by the fact that a more robust and flexible tracker
can be attained if a history of variations in the object appearance is kept~\cite{kalal2011tracking}.
To understand the benefit of the bag of models,
assume a person tracking is desired where the appearance of whole body is encoded as an object model.
Moreover, assume at some point of time only the upper body of person is visible (due to partial occlusion)
and the tracker has successfully learned the new appearance.
If the tracking system is only aware of the very last seen appearance (upper-body in our example),
upon termination of occlusion,
the tracker is likely to lose the object.
Keeping a set of models (in our example both upper-body and whole body) can help the tracking system to cope with drastic changes.

\end{enumerate}

\noindent
Each of the components is elucidated in the following subsections.

\subsection{Motion Estimation}
\label{Motion_estimation}

In the proposed framework, we are aiming to obtain the location {\small $x \in \mathcal{X}$, $y \in \mathcal{Y}$}
and the scale {\small $s \in \mathcal{S}$} of an object in frame $t$ based on prior knowledge about previous frames.
A~blind search in the space of {\small $\mathcal{X}-\mathcal{Y}-\mathcal{S}$} is obviously inefficient,
since not all possible combinations of $x$, $y$ and $s$ are plausible.
To efficiently search the {\small $\mathcal{X}-\mathcal{Y}-\mathcal{S}$} space,
we use a sequential Monte Carlo method known as the Condensation algorithm~\cite{isard1996contour}
to determine which combinations in the {\small $\mathcal{X}-\mathcal{Y}-\mathcal{S}$} space are most probable at time $t$.
The key idea is to represent the {\small $\mathcal{X}-\mathcal{Y}-\mathcal{S}$} space by a density function
and estimate it through a set of random samples (also known as particles).
As the number of particles becomes large, the condensation method approaches
the optimal Bayesian estimate of density function (\ie combinations in the {\small $\mathcal{X}-\mathcal{Y}-\mathcal{S}$} space).
Below, we briefly describe how the condensation algorithm is used within the proposed tracking approach.

Let {\small $\mathcal{Z}^{(t)} = \left(x^{(t)},y^{(t)},s^{(t)}\right)$} denote a particle at time $t$.
By the virtue of the principle of importance sampling~\cite{maskell2001tutorial},
the density of {\small $\mathcal{X}-\mathcal{Y}-\mathcal{S}$} space (or most probable candidates) at time $t$ is estimated as a set of $N$ particles
{\small $\{\mathcal{Z}_{i}^{(t)}\}_{i=1}^{N}$} using previous particles {\small $\{\mathcal{Z}^{(t-1)}_{i}\}_{i=1}^{N}$} and
their associated weights \mbox{\small{$\{w^{(t-1)}_{i}\}_{i=1}^{N}$}}
with {\small $\sum\nolimits_{i=1}^N w^{(t-1)}_{i} = 1$}.
For now we assume the associated weights of particles are known and later discuss how they can be determined.

In the condensation algorithm, to generate {\small $\{\mathcal{Z}_{i}^{(t)}\}_{i=1}^{N}$},
{\small $\{\mathcal{Z}^{(t-1)}_{i}\}_{i=1}^{N}$} is first sampled (with replacement) $N$ times.
The probability of choosing a given element {\small $\mathcal{Z}^{(t-1)}_{i}$} is equal to the associated weight {\small $w^{(t-1)}_{i}$}.
Therefore, the particles with high weights might be selected several times, leading to identical
copies of elements in the new set. Others with relatively low weights may not be chosen at all.
Next, each chosen element undergoes an independent Brownian motion step.
Here, the Brownian motion of a particle is modelled by a Gaussian distribution with a diagonal covariance matrix.
As a result, for a chosen particle {\small $\mathcal{Z}^{(t-1)}_{\ast}$} from the first step of condensation algorithm,
a new particle {\small $\mathcal{Z}^{(t)}_{\ast}$}
is obtained as a random sample of {\small $\mathcal{N}\left(\mathcal{Z}^{(t-1)}_*,\Sigma\right)$}
where {\small $\mathcal{N}\left(\mu,\Sigma\right)$}
denotes a Gaussian distribution with mean {\small $\mu$} and covariance {\small $\Sigma$}.
The covariance {\small $\Sigma$} governs the speed of motion, and is a constant parameter over time in our framework.

\subsection{Candidate Templates}
\label{Candidate_templates}

\begin{figure}[!b]
  \centerline{\includegraphics[width=1\columnwidth,keepaspectratio]{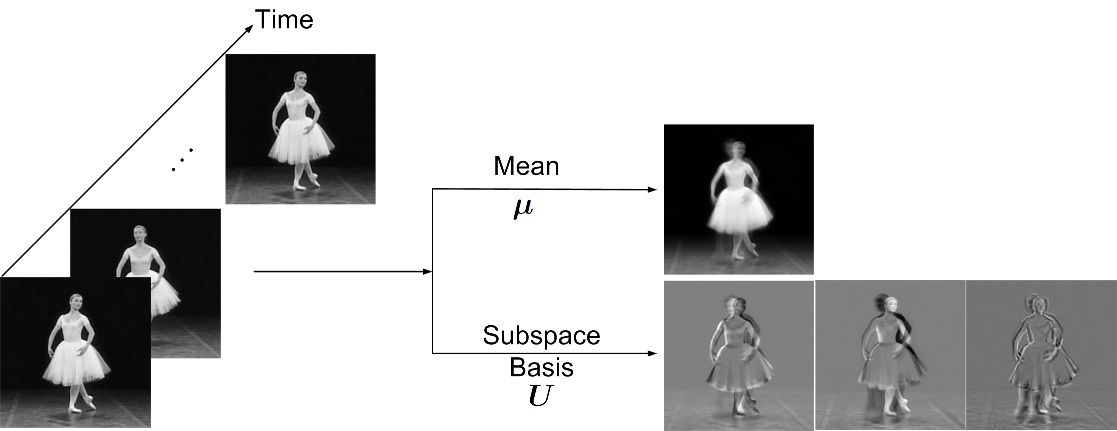}}
  \caption
    {
    \small
    In the proposed approach, object appearance is modelled by an affine subspace. An affine subspace is uniquely described by its origin $\mu$ and basis $U$. Here, $\mu$ and basis $U$ are obtained by computing mean and eigenbasis of a set of object images.
    }
  \label{fig:app_mod}
\end{figure}

To accommodate variations in object appearance,  this module models the appearance of particles%
\footnote{We loosely use ``particle appearance'' to mean the appearance of a candidate template described by a particle.}
by affine subspaces
(see Fig.~\ref{fig:app_mod} for a conceptual example).
An affine subspace is a subset of Euclidean space~\cite{von2007tutorial},
formally described by a 2-tuple \{$\mu,U$\} as:

\noindent
\begin{small}
\begin{equation}
    \Mat{A} = \left \{ \Mat{z} \in \mathbb{R}^{D}: \Mat{z} = \mu + \Mat{U} \Vec{y} \right \}
    \label{eqn:raff}
\end{equation}%
\end{small}%

\noindent
where
{\small $\mu \in \mathbb{R}^{D}$}
and
{\small $\Mat{U} \in \mathbb{R}^{D \times n}$}
are origin and basis of the subspace, respectively.
Let {\small $\Mat{I}(\mathcal{Z}^{(t)}_{\ast},t)$} denote the vector representation of an {\small $N_{1} \times N_{2}$} patch
extracted from frame $t$ by considering the values of particle {\small $\mathcal{Z}^{(t)}_{\ast}$}.
That is, frame $t$ is first scaled appropriately based on the value \mbox{\small{$s^{(t)}_\ast$}}
and then a patch of {\small $N_{1} \times N_{2}$} pixels with the top left corner located at
{\small $\left(x^{(t)}_\ast,y^{(t)}_\ast\right)$} is extracted.

The appearance model for {\small $\mathcal{Z}^{(t)}_{\ast}$} is generated from
a set of {\small $P+1$} images by considering $P$ previous results of tracking.
More specifically, let {\small $\widehat{\mathcal{Z}}^{(t)}$} denote the result of tracking at time $t$,
\ie~{\small $\widehat{\mathcal{Z}}^{(t)}$} is the most similar particle to the bag of models at time $t$.
Then set
\mbox{\footnotesize $
\mathbb{B}_{\mathcal{Z}_{\ast}}^{(t)} \mbox{~=~}
\left \{\Mat{I}(\widehat{\mathcal{Z}}^{(t-P)},t \mbox{~--~} P),
\Mat{I}(\widehat{\mathcal{Z}}^{(t-P+1)},t \mbox{~--~} P+1),\cdots, \Mat{I}(\mathcal{Z}^{(t)}_{\ast},t) \right \}$}
is used to obtain the appearance model for particle {\small $\mathcal{Z}^{(t)}_{\ast}$}.
More specifically, the origin of affine subspace associated to {\small $\mathcal{Z}^{(t)}_{\ast}$}
is the mean of {\small $\mathbb{B}_{\mathcal{Z}_{\ast}}^{(t)}$}.
The basis is obtained by computing the Singular Value Decomposition (SVD) of
{\small $\mathbb{B}_{\mathcal{Z}_{\ast}}^{(t)}$} and choosing the $n$ dominant left-singular vectors.

\subsection{Bag of Models}
\label{Bag_templates}

Although affine subspaces accommodate object changes along with a set of images,
to produce a robust tracker,
the object's model should be able to reflect the appearance changes during the tracking process.
Accordingly, we propose to keep a set of object models {\small $m_{j}=\{\mu_{j},U_{j}\}$}
for coping with deformations, pose variations, occlusions, and other variations of the object during tracking.

\begin{figure}[!b]
  \begin{minipage}{1\columnwidth}
    \begin{minipage}{0.05\columnwidth}
      \centerline{\bf (a)}
    \end{minipage}
    \begin{minipage}{0.95\columnwidth}
      \includegraphics[width=1\textwidth,keepaspectratio]{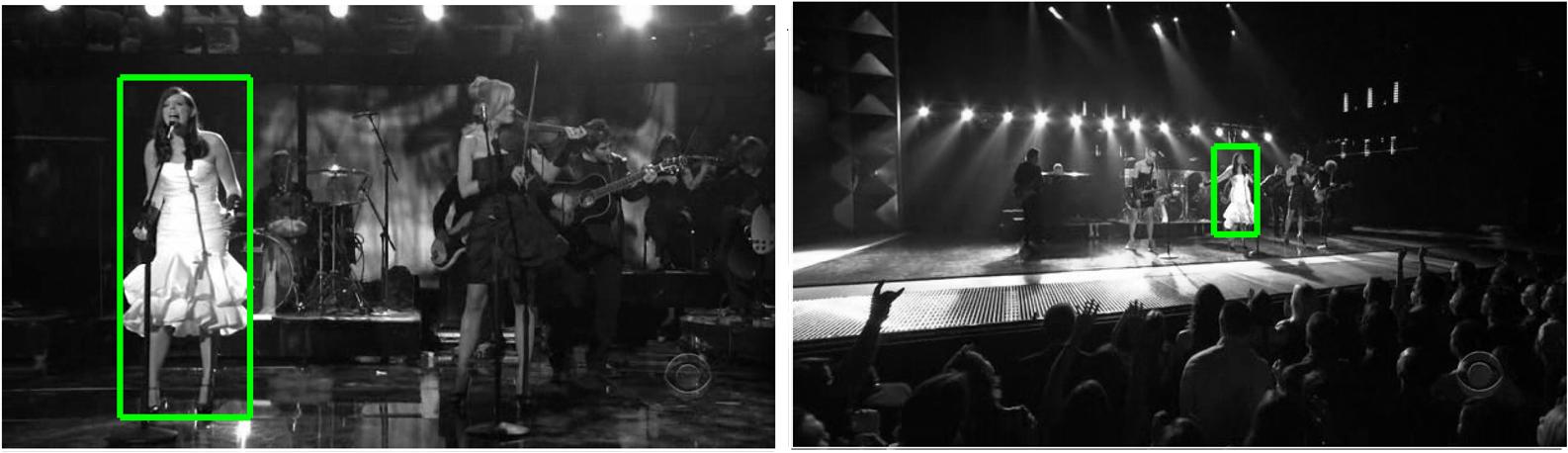}
    \end{minipage}
  \end{minipage}
  \begin{minipage}{1\columnwidth}
    \begin{minipage}{0.05\columnwidth}
      \centerline{\bf (b)}
    \end{minipage}
    \begin{minipage}{0.95\columnwidth}
      \includegraphics[width=1\textwidth,keepaspectratio]{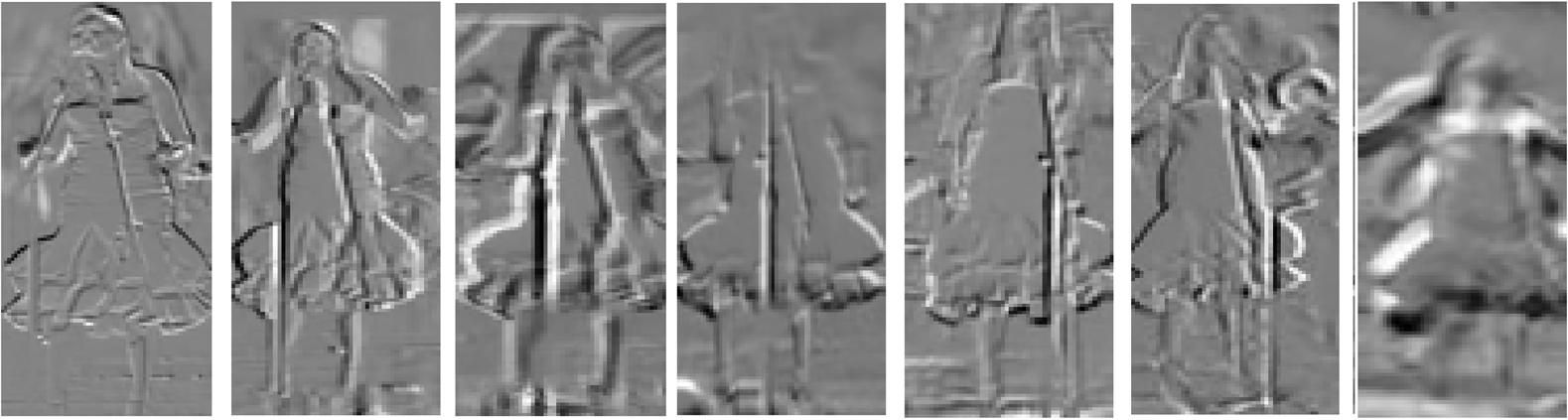}
    \end{minipage}
  \end{minipage}

\begin{minipage}{1\columnwidth}
    \begin{minipage}{0.05\columnwidth}
      \centerline{\bf (c)}
    \end{minipage}
    \begin{minipage}{0.95\columnwidth}
      \includegraphics[width=1\textwidth,keepaspectratio]{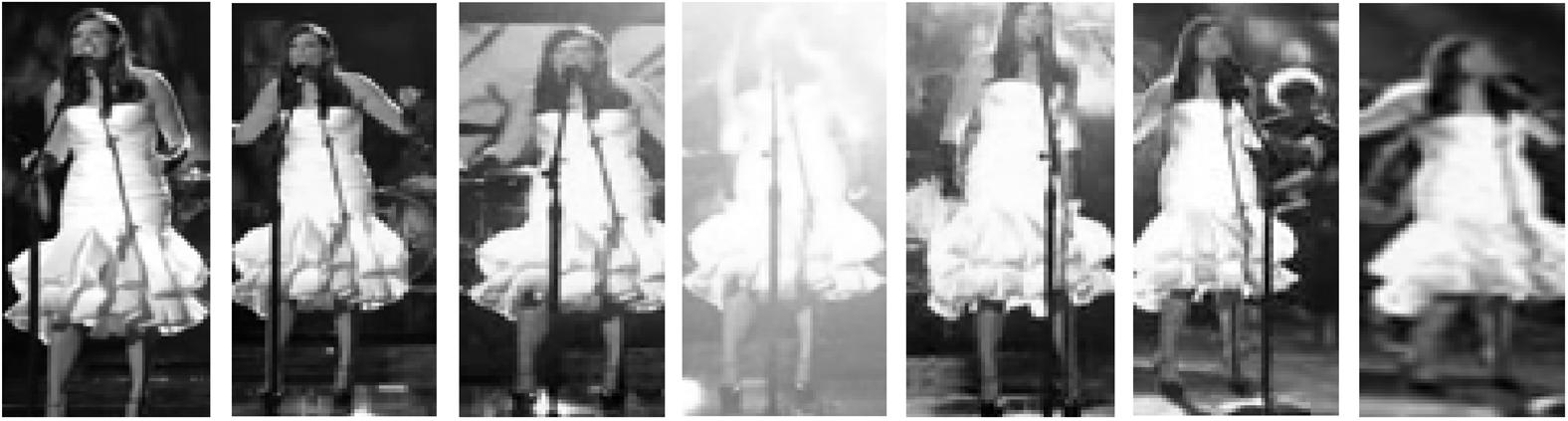}
    \end{minipage}
  \end{minipage}
  \vspace{1ex}
  \caption
    {
    {\bf (a)}~Two examples of a frame with a tracked object.
    {\bf (b)}~The first eigenbasis of ten sample template bags.
    {\bf (c)}~The recent frame in each of the 10 image sets used to generate the templates.
    }
  \label{fig:tmp_example}
\end{figure}

Fig.~\ref{fig:tmp_example} shows two frames with a tracked object,
the bag models used to localise the object,
and the recent images of the image set used to generate each bag model.

\begin{figure}[!b]
  \centerline{\includegraphics[width=1.025\columnwidth,keepaspectratio]{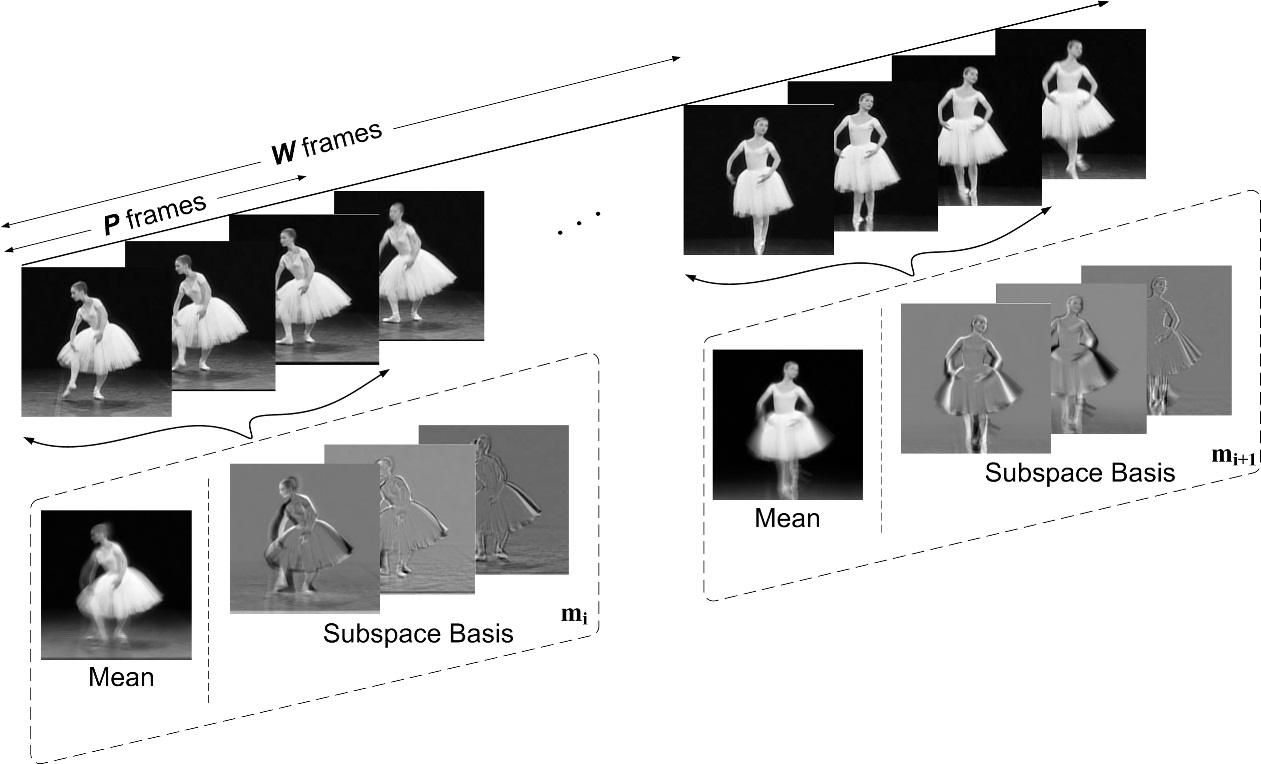}}
  \caption
    {
    \small
    The model extraction procedure involves a sliding window update scheme.
    The template is learned from a set of \textit{P} consecutive frames.
    Template update occurs every \textit{W} frames.
    }
  \label{fig:template}
\end{figure}

A bag {\small $\mathcal{M} = \{m_{1},\cdots,m_{k}\}$} is defined as a set of $k$ object models,
\ie~each $m_j$ is an affine subspace learned during the tracking process.
The bag is updated every $W$ frames (see Fig.~\ref{fig:template}) by replacing the oldest model with the latest learned model
(\ie~latest result of tracking specified by {\small $\widehat{\mathcal{Z}}^{(t)}$}).
The size of bag $k$ determines the memory of the tracking system.
Thus, a large bag with several models might be required to track an object in a challenging scenario.
In our experiments, a bag of size $10$ with the updating rate $W = 5$ is used in all experiments.

Having a set of models at our disposal, we will next address how the similarity between a particle's appearance and the bag can be determined.

\subsection{Decision Making}
\label{decision_making}

Given the previously learned affine subspaces as the input to this module, the aim is to find the nearest affine subspace to the bag templates.
Although the minimal Euclidean distance is the simplest distance measure between two affine subspaces
(\ie the minimum distance of any pair of points of the two subspaces),
this measure does not form a metric \cite{basri2011approximate}
and it does not consider the angular distance between affine subspaces, which can be a useful discriminator~\cite{kim2007discriminative}.
However, the angular distance ignores the origin of affine subspaces and simplifies the problem to a linear subspace case,
which we wish to avoid.

To address the above limitations, we propose a distance measure with the following form:

\vspace{-1ex}
\noindent
\begin{small}
\begin{equation}
    \operatorname{dist}(\Mat{A}_{i},\Mat{A}_{j})
    \mbox{~=~}
    \operatorname{dist}_G\left(\Mat{U}_{i},\Mat{U}_{j}\right)
    +
    \alpha  (\Vec{\mu}_{i} - \Vec{\mu}_{j})^T\Mat{M}(\Vec{\mu}_{i}-\Vec{\mu}_{j})
    \label{eqn:r2}
\end{equation}%
\end{small}%

\noindent
where $\operatorname{dist}_G$ is the Geodesic distance between two points on a Grassmann manifold~\cite{edelman1998geometry}, \mbox{\small{$(\Vec{\mu}_{i}-\Vec{\mu}_{j})^T\Mat{M}(\Vec{\mu}_{i}-\Vec{\mu}_{j})$}} is the Mahalanobis
distance between origins of  $\Mat{A}_{i}$ and $\Mat{A}_{j}$,
and $\alpha$ is a mixing weight.
The components in the proposed distance are described below.

A Grassmann manifold (a special type of Riemannian manifold)
is defined as the space of all $n$-dimensional linear subspaces of {\small $\mathbb{R}^D$} for \mbox{\small $0<n<D$}.
A point on Grassmann manifold {\small $\mathcal{G}_{D,n}$}
is represented by an orthonormal basis through a \mbox{\small $D \times n$} matrix.
The length of the shortest smooth curve connecting two points on a manifold is known as the geodesic distance.
For Grassmann manifolds, the geodesic distance is given by:

\vspace{-2ex}
\noindent
\begin{small}
\begin{equation}
    \operatorname{dist}_G\left(\Mat{X},\Mat{Y}\right)=\|\Theta\|_2
    \label{eqn:geodesic_Grass}
\end{equation}%
\end{small}%

\noindent
where {\small $\Theta=[\theta_1,\theta_2,\cdots,\theta_n]$} is the principal angle vector, \ie

\vspace{-1ex}
\noindent
\begin{small}
\begin{equation}
  \cos(\theta_l)
  =
  \max_{\Vec{x} \in \Mat{X},\Vec{y} \in \Mat{Y}}
  \Vec{x}^T \Vec{y} = \Vec{x}_l^T\Vec{y}_l
  \label{eqn:Principal_Angle}
\end{equation}%
\end{small}%

\noindent
subject to
\mbox{\small $\left \| \Vec{x} \right \| \mbox{~=~} \left \| \Vec{x} \right \| \mbox{~=~} 1$},
\mbox{\small $\Vec{x}^T \Vec{x}_i \mbox{~=~} \Vec{y}^T \Vec{y}_i \mbox{~=~} 0$},
\mbox{\small $i \mbox{~=~} 1, \ldots, l-1$}.
The principal angles have the property of \mbox{$\theta_i \in [0, \pi/2]$}
and can be computed through the SVD of \mbox{\small {$\Mat{X}^T \Mat{Y}$}}~\cite{edelman1998geometry}.

We note that the linear combination of a Grassmann distance (distance between linear subspaces)
and Mahalanobis distance (between origins) of two affine subspaces has roots in probabilistic subspace distances~\cite{hamm2009extended}.
More specifically, consider two normal distributions
{\small $\mathcal{N}_1\left(\Vec{\mu}_1,\Mat{C}_1\right)$}
and
{\small $\mathcal{N}_2\left(\Vec{\mu}_2,\Mat{C}_2\right)$}
with \mbox{\small {$\Mat{C}_{i} = \sigma^{2}\mathbb{I} + \Mat{U}_{i}\Mat{U}_{i}^T$}} as the covariance matrix, and $\Vec{\mu}_{i}$ as the mean vector.
The symmetric Kullback-Leibler (KL) distance between $\mathcal{N}_1$ and $\mathcal{N}_2$ under
orthonormality condition (\ie \mbox{\small{$\Mat{U}_{i}^T \Mat{U}_{i}=\mathbb{I}_{n}$}}) results in:

\vspace{-3ex}
\noindent
\begin{small}
\begin{eqnarray}
    J_{KL} & \mbox{=} &\frac{1}{2\sigma^2}(\Vec{\mu}_{1} \mbox{--} \Vec{\mu}_{2})^T\left(2\mathbb{I}_{D} \mbox{~--~} \Mat{U}_{1}\Mat{U}_{1}^T  \mbox{~--~} \Mat{U}_{2} \Mat{U}_{2}^T\right)(\Vec{\mu}_{1}  \mbox{--} \Vec{\mu}_{2}) \nonumber \\
    & & + ~ \frac{1}{2\sigma^2(\sigma^2+1)}\left(2n - 2\tr(\Mat{U}_{1}^T \Mat{U}_{2}\Mat{U}_{2}^T \Mat{U}_{1})\right)
    \label{eqn:r3}
\end{eqnarray}%
\end{small}%

The term {\small $\tr(\Mat{U}_{1}^T \Mat{U}_{2}\Mat{U}_{2}^T \Mat{U}_{1})$ in $J_{KL}$}
is identified as the projection distance on Grassmann manifold {\small $\mathcal{G}_{D,n}$}
(defined as {\small $\operatorname{dist}_{Proj}\left(\Mat{U}_1,\Mat{U}_2\right)=\|sin(\Theta)\|_2$})~\cite{hamm2009extended},
and the term
\mbox{\small $(\Vec{\mu}_{1}-\Vec{\mu}_{2})^T\left(2\mathbb{I}_{D} - \Mat{U}_{1}\Mat{U}_{1}^T - \Mat{U}_{2} \Mat{U}_{2}^T\right)(\Vec{\mu}_{1}-\Vec{\mu}_{2})$}
is the Mahalanobis distance with
{\small $\Mat{M} = 2\mathbb{I}_{D} - \Mat{U}_{1}\Mat{U}_{1}^T - \Mat{U}_{2} \Mat{U}_{2}^T$}.

Since the geodesic distance is a more natural choice for measuring lengths on Grassmann manifolds
(compared to the projection distance),
we have elected to combine it with the Mahalanobis distance from (\ref{eqn:r3}),
resulting in the following instantiation of the general form given in Eqn.~(\ref{eqn:r2}):

\noindent
\begin{footnotesize}
\begin{eqnarray*}
    \operatorname{dist}(\Mat{A}_{i},\Mat{A}_{j}) \hspace{-1.5ex}
    & \mbox{=} &
    \hspace{-1.5ex} \operatorname{dist}_{G}(\Mat{U}_{i},\Mat{U}_{j})\nonumber \\
    & &
    \hspace{-1.5ex} + ~
    \alpha
    (\Vec{\mu}_{i} \mbox{~--~} \Vec{\mu}_{j})^T
    \left(2\mathbb{I}_{D} \mbox{~--~} \Mat{U}_{i}\Mat{U}_{i}^T \mbox{~--~} \Mat{U}_{j} \Mat{U}_{j}^T\right)
    (\Vec{\mu}_{i} \mbox{~--~} \Vec{\mu}_{j})
\end{eqnarray*}%
\end{footnotesize}%

~

We measure the likelihood of a candidate subspace $A_{i}^{(t)}$,
given template $m_{j}$,
as follows:

\noindent
\begin{small}
\begin{equation}
    p \left(  A_{i}^{(t)}|m_{j} \right) = \exp\left(\frac{-\operatorname{dist}(A_{i}^{(t)},m_{j})}{\sigma}\right) 
    \label{eqn:r6}
\end{equation}%
\end{small}%

\noindent
where $\sigma$ indicates the standard deviation of the likelihood function and is a parameter in the tracking framework.
The likelihoods are normalised such that {\small $\sum_{i=1}^{N}p(A_{i}^{(t)}|m_{j})=1$}.
To measure the likelihood between a candidate affine subspace {\small $A_{i}^{(t)}$} and bag {\small $\mathcal{M}$},
the individual likelihoods between {\small $A_{i}^{(t)}$} and bag templates {\small $m_{j}$} should be integrated.
Based on~\cite{kittler1998combining}, we opt for the sum rule:

\noindent
\begin{small}
\begin{equation}
    p(A_{i}^{(t)}|\mathcal{M}) = \sum\nolimits_j^k{p(A_{i}^{(t)}|m_{j})}
    \label{eqn:r8}
\end{equation}%
\end{small}%

\noindent
The object state is then estimated as:

\noindent
\begin{small}
\begin{equation}
    \widehat{\mathcal{Z}}^{(t)} = \mathcal{Z}_j^{(t)},~~~~~\mbox{where}~~~j = \underset{i}{\operatorname{argmax}}~~p(A_{i}^{(t)}|\mathcal{M})
    \label{eqn:r9}
\end{equation}%
\end{small}%


\begin{algorithm}[!b]
\caption{: Affine Subspace Tracking}
\small
\label{alg:tracking_pseudo_code}
\begin{algorithmic}[1]

\REQUIRE
~\\
\begin{itemize}
\vspace{-2ex}
\item
New frame, a set of updated candidate object states from the last frame,
and the previous $\small{P-1}$ estimated object states $\{\widehat{\mathcal{Z}}^{(\tau)}\}_{\tau = t-P+1}^{t-1}$
\end{itemize}
\vspace{0.25ex}
\STATE {\bf Initialisation:}
~\\
\begin{itemize}
\vspace{-1ex}
\item
$t = 1:P$
\vspace{-1ex}
\item
Set the initial object state $\widehat{\mathcal{Z}}^{(t)}$ in the first $\small P$ frames.

\item
Use a single state to indicate the location.

\end{itemize}

\STATE {\bf Begin:}
\begin{itemize}
\item
Select candidate object states according to the dynamic model $\{\mathcal{Z}_{i}^{(t)}\}_{i=1}^{N}$

\item
For each sample, extract the corresponding image patch

\item
For each $\mathcal{Z}^{(t)}_{i}$ do:
\vspace{-1ex}
\begin{itemize}
\item
Generate the affine subspace $A_{i}^{(t)}\{\mu_{i}^{(t)},U_{i}^{(t)}\}$ based on image regions corresponding to $\mathcal{Z}^{(t)}_{i}$ and $\{\widehat{\mathcal{Z}}^{(\tau)}\}_{\tau = t-P+1}^{t-1}$

\item
Calculate the likelihoods given each template in the bag by Eqn.~(\ref{eqn:r6})

\item
Compute the final likelihoods using Eqn.~(\ref{eqn:r8})

\end{itemize}
\item
Determine the object state $\widehat{\mathcal{Z}}^{(t)}$ by Maximum Likelihood (ML) estimation

\item
Update the existing candidate object states according to their probabilities~\cite{isard1996contour}

\end{itemize}

\ENSURE
current object state $\widehat{\mathcal{Z}}^{(t)}$

\end{algorithmic}
\end{algorithm}

\subsection{Computational Complexity}
\label{Comp_complexity}

The computational complexity of the proposed tracking framework
can be associated with generating a new model and comparing a target candidate with a model.
The model generation step requires {\small $O(D^3+2Dn)$} operations.
Computing the geodesic distance between two points on $G_{D,n}$ requires {\small \mbox{$O((D+1)n^2+n^3)$}} operations.
Therefore, comparing an affine subspace candidate against each bag template needs {\small \mbox{ $O((2n+3)D^2+(n^2+1)D+n^3+n^2)$}} operations. 

\section{Experiments}
\label{sec:experiments_v4}

\begin{figure*}[!tb]
  \begin{minipage}{1\textwidth}
    \begin{minipage}{0.05\textwidth}
      \centerline{\bf (a)}
    \end{minipage}
    \begin{minipage}{0.95\textwidth}
      \includegraphics[width=1\textwidth,keepaspectratio]{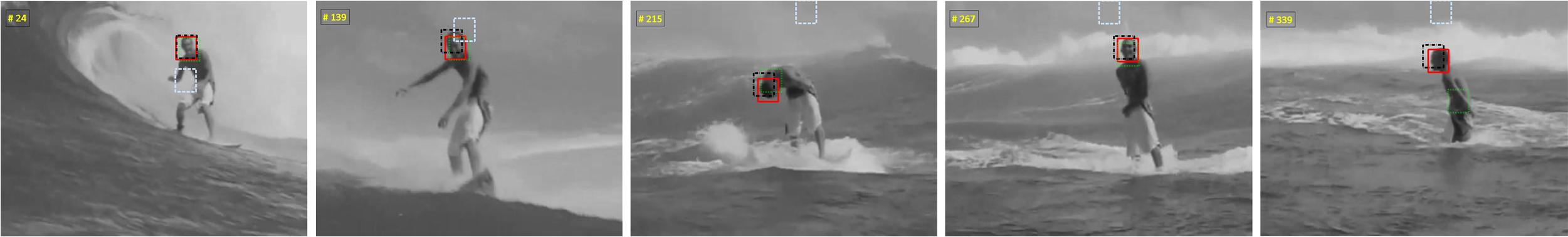}
    \end{minipage}
  \end{minipage}
  \begin{minipage}{1\textwidth}
    \begin{minipage}{0.05\textwidth}
      \centerline{\bf (b)}
    \end{minipage}
    \begin{minipage}{0.95\textwidth}
      \includegraphics[width=1\textwidth,keepaspectratio]{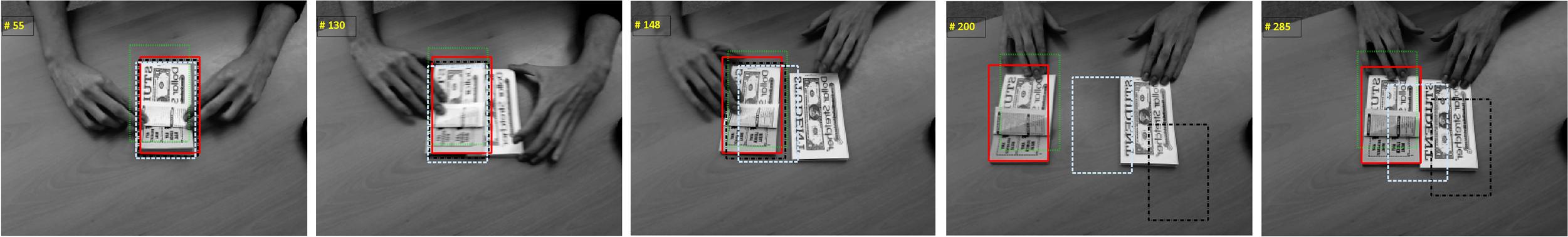}
    \end{minipage}
  \end{minipage}
  \begin{minipage}{1\textwidth}
    \begin{minipage}{0.05\textwidth}
      \centerline{\bf (c)}
    \end{minipage}
    \begin{minipage}{0.95\textwidth}
      \includegraphics[width=1\textwidth,keepaspectratio]{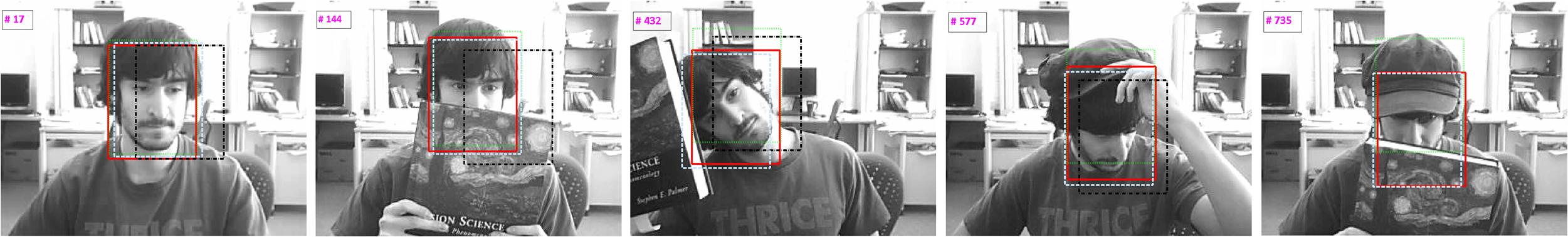}
    \end{minipage}
  \end{minipage}
  \begin{minipage}{1\textwidth}
    \begin{minipage}{0.05\textwidth}
      \centerline{\bf (d)}
    \end{minipage}
    \begin{minipage}{0.95\textwidth}
      \includegraphics[width=1\textwidth,keepaspectratio]{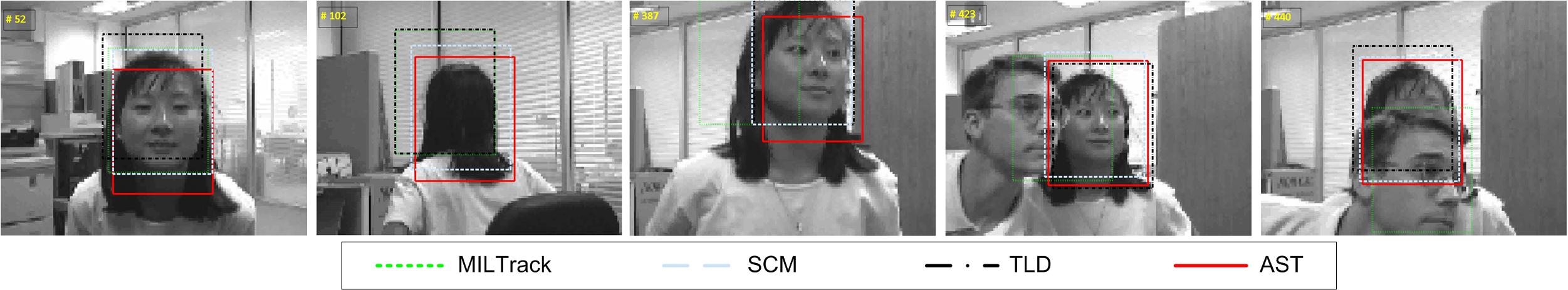}
    \end{minipage}
  \end{minipage}

  \vspace{1ex}
  \caption
    {
    Examples of bounding boxes resulting from tracking on several video sequences.
    For the sake of clarity, we only demonstrate the results of the overall top four trackers.
    {\bf (a)}~{\it Surfer}~\cite{babenko2011}: includes large pose variations, occlusion;
    {\bf (b)}~{\it Coupon Book}~\cite{babenko2011}: contains severe appearance change in addition to including an imposter to distract the tracker;
    {\bf (c)}~{\it Occluded Face~2}~\cite{babenko2011}: contains various occlusions;
    {\bf (d)}~{\it Girl}~\cite{birchfield1998elliptical} involves partial and full occlusion, large pose changes.
    }
  \label{fig:scr_sh3}
  \vspace{0.5ex}
  \hrule
  \vspace{-1ex}
\end{figure*}

In this section we evaluate and analyse the performance of the proposed AST method using eight publicly available videos%
\footnote[6]{The videos and the corresponding ground truth are available at {\it http://vision.ucsd.edu/\~{}bbabenko/project\_miltrack.shtml}}
consisting of two main tracking tasks: face and object tracking.
The sequences are:
{\it Occluded Face}~\cite{adam2006robust},
{\it Occluded Face~2}~\cite{babenko2011},
{\it Girl}~\cite{birchfield1998elliptical},
{\it Tiger~1}~\cite{babenko2011},
{\it Tiger~2}~\cite{babenko2011},
{\it Coke~Can}~\cite{babenko2011},
{\it Surfer}~\cite{babenko2011},
and
{\it Coupon~Book}~\cite{babenko2011}.
Example frames from several videos are shown in Fig.~\ref{fig:scr_sh3}.

Each video is composed of 8-bit grayscale images, resized to {\small $320 \times 240$} pixels.
We used the raw pixel values as image features.
For the sake of computational efficiency in the affine subspace representation,
we resized each candidate image region to {\small $32 \times 32$}, and the number of eigenvectors ($n$) used in all experiments is set to three.
Furthermore, we only consider 2D translation and scaling in the motion modelling component.
The batch size ({\small $W$}) for the template update is set to five
as a trade-off between computational efficiency and effectiveness of modelling appearance change during fast motion.

We evaluated the proposed tracker based on {\bf (i)}~average center location error, and {\bf (ii)}~precision~\cite{babenko2011}.
Precision shows the percentage of frames for which the estimated object location is within a threshold distance of the ground truth.
Following~\cite{babenko2011}, we use a fixed threshold of 20 pixels.

To contrast the effect of affine subspace modelling against linear subspaces,
we assessed the performance of the AST tracker against a tracker that only exploits linear subspaces,
\ie, an AST where $\mu = 0$ for all models.
The results, in terms of center location errors, are shown in \mbox{Table~\ref{tab:linear}}.
The proposed AST method significantly outperforms the linear subspaces approach,
thereby confirming our idea of affine subspace modelling.

\begin{table}[!tb]
  \centering
  ~
  \caption
    {
    \small
    Performance comparison between tracking based on affine and linear subspaces,
    in terms of average center location errors (pixels).
    }
  \label{tab:linear}

  ~  

  \scalebox{1}{
  \begin{tabular}{lcc}
    \toprule
    \multirow{1}{*} {\bf Video}    &{\bf proposed AST}  &{\bf linear subspace}\\
    \toprule
    {Surfer}        &${8}$     &$39$     \\
    {Coke Can}        &${9}$     &$31$    \\
    {Girl}             &${19}$     &$29$    \\
    {Tiger 1}        &${22}$     &$38$   \\
    {Tiger 2}        &${15}$     &$42$    \\
    {Coupon Book}        &${8}$     &$25$\\
    {Occluded Face}       & ${14}$      & $27$            \\
    {Occluded Face 2}             & $ {13}$      & $24$            \\
    \bottomrule
    {\bf average error}    &${\bf13.5}$ & $\bf31.88$\\
    \bottomrule
  \end{tabular}}
\end{table}

~

\newpage
\subsection{Quantitative Comparison}
\vspace{-1ex}

To assess and contrast the performance of AST tracker against
state-of-the-art methods, we consider six methods, here.
The competitors are:
fragment-based tracker (FragTrack)~\cite{adam2006robust},
multiple instance boosting-based tracker (MILTrack)~\cite{babenko2011,babenko2009visual},
online Adaboost (OAB)~\cite{grabner2006real},
tracking-learning-detection (TLD)~\cite{kalal2011tracking},
incremental visual tracking (IVT)~\cite{ross2008incremental}, and Sparsity-based Collaborative Model tracker (SCM)~\cite{zhong2012}.
We use the publicly available source codes for FragTrack\footnote[1]{{\it http://www.cs.technion.ac.il/\~{}amita/fragtrack/fragtrack.htm}},
MILTrack\footnote[2]{{\it http://vision.ucsd.edu/~bbabenko/project\_miltrack.shtml}},
OAB\footnotemark[2],
TLD\footnote[3]{{\it http://info.ee.surrey.ac.uk/Personal/Z.Kalal/}},
IVT\footnote[4]{{\it http://www.cs.toronto.edu/\~{}dross/ivt/}}
and SCM\footnote[5]{{\it http://ice.dlut.edu.cn/lu/Project/cvpr12\_scm/cvpr12\_scm.htm}}.

Tables~\ref{tab:errors} and~\ref{tab:precision} show the performance in terms of precision and location error, respectively,
for the proposed AST method as well as the competing trackers.
Fig.~\ref{fig:scr_sh3} shows resulting bounding boxes for several frames
from the {\it Surfer}, {\it Coupon Book}, {\it Occluded Face~2} and {\it Girl} sequences.
On average, the proposed AST method obtains notably better performance than the competing trackers,
with TLD being the second best tracker.

\begin{table*}[!tb]
  \centering
  \begin{minipage}{0.49\textwidth}
    \centering
    \caption
      {
      \small
      Comparison of the proposed AST method against competing trackers,
      in terms of average center location errors (pixels).
      Best performance is indicated by $\ast$,
      while second best by $\ast\ast$.
      }
    \label{tab:errors}
    \vspace{1ex}
    \scalebox{0.62}
    {
      \begin{tabular}{llllllllll}
        \toprule
        \multirow{2}{*}{\hspace{-1ex}\bf Video}    &\hspace{-1ex}{\bf AST}  &\hspace{-1ex}{\bf TLD}          &\hspace{-1ex}{\bf MILTrack}            &\hspace{-1ex}{\bf SCM}  &\hspace{-1ex}{\bf OAB} &\hspace{-1ex}{\bf IVT} &\hspace{-1ex}{\bf FragTrack}\\
                                      &\hspace{-1ex}(proposed) &\hspace{-1ex}{\bf\cite{kalal2011tracking}} &\hspace{-1ex}{\bf\cite{babenko2011}} &\hspace{-1ex}{\bf\cite{zhong2012}} &\hspace{-1ex}{\bf\cite{grabner2006real}} &\hspace{-1ex}{\bf\cite{ross2008incremental}}
                                      &\hspace{-1ex}{\bf\cite{adam2006robust}}\\
        \toprule
        \hspace{-1ex}{Surfer}              &${~~8~{\ast}}$     &$~~{9~{\ast\ast}}$ &$11$     &${76}$     &$23$    &$30$  &$139$\\
        \hspace{-1ex}{Coke Can}            &${~~9~{\ast}}$   &${13~{\ast\ast}}$   &$20$      &${~~9~{\ast}}$      &$25$    &$61$ &$~~63$\\
        \hspace{-1ex}{Girl}                &${19~{\ast\ast}}$  &$28$     &$32$      &${10~{\ast}}$     &$48$      &$52$     &$~~{27}$\\
        \hspace{-1ex}{Tiger 1}             &$22$        &${10~\ast}$        &${16~{\ast\ast}}$  &${37}$ &$35$      &$59$ &$~~39$\\
        \hspace{-1ex}{Tiger 2}             &${15~{\ast}}$    &${15~{\ast}}$   &${18~{\ast\ast}}$  &${43}$ &$33$      &$43$ &$~~37$\\
        \hspace{-1ex}{Coupon Book}         &${~~8~{\ast}}$    &$37$  &${15~{\ast\ast}}$   &${36}$ &$25$      &$17$ &$~~56$\\
        \hspace{-1ex}{Occluded Face}       &${14~~}$      & ${16}$      & $27$      &${~~4~{\ast}}$    &$43$      &${~9}$  &${~~~~6~{\ast\ast}}~$\\
        \hspace{-1ex}{Occluded Face 2}     &${13~{\ast\ast}}$   &$28$    & ${20}$ &${~~8~{\ast}}$ &$21$      &${17}$ &$~~45$           \\
        \bottomrule
        \hspace{-1ex}{\bf average error}   &${\bf13.5~{\ast}}$ &${\bf19.49~{\ast\ast}}$ &$\bf19.87$ &${\bf27.87}$ &$\bf 31.62$ &$\bf36.00$ &$\bf 51.5$\\
        \bottomrule
      \end{tabular}
    }
  \end{minipage}
  \hfill
  \begin{minipage}{0.49\textwidth}
    \centering
    \caption
      {
      \small
      Precision at a fixed threshold of 20, as per~\cite{babenko2011}.
      Best performance is indicated by~$\ast$,
      while second best is indicated by~$\ast\ast$.
      The higher the precision, the better.
      }
    \label{tab:precision}
    \vspace{1ex}
    \scalebox{0.62}{
    \vspace{1ex}

    \begin{tabular}{lllllllllllll}
      \toprule
      \multirow{2}{*}{\hspace{-1ex}\bf Video}    &\hspace{-1ex}{\bf AST}  &\hspace{-1ex}{\bf TLD}          &\hspace{-1ex}{\bf MILTrack}            &\hspace{-1ex}{\bf SCM}  &\hspace{-1ex}{\bf OAB} &\hspace{-1ex}{\bf IVT} &\hspace{-1ex}{\bf FragTrack}\\
                                    &\hspace{-1ex}(proposed) &\hspace{-1ex}{\bf\cite{kalal2011tracking}} &\hspace{-1ex}{\bf\cite{babenko2011}} &\hspace{-1ex}{\bf\cite{zhong2012}} &\hspace{-1ex}{\bf\cite{grabner2006real}} &\hspace{-1ex}{\bf\cite{ross2008incremental}}
                                    &\hspace{-1ex}{\bf\cite{adam2006robust}}\\
      \toprule
      \hspace{-1ex}{Surfer}        &${0.98~{\ast}}$   &${0.97~{\ast\ast}}$  &$0.93$   &${0.10}$  &$0.51$      &$0.19$ &$0.28$\\
      \hspace{-1ex}{Coke Can}        &${0.99~{\ast}}$   &${0.98~{\ast\ast}}$  &$0.55$   &${0.97}$ &$0.45$      &$0.13$ &$0.14$\\
      \hspace{-1ex}{Girl}             &${0.73~{\ast\ast}}$  &$0.42$   &$0.32$   &${0.97~{\ast}}$ &$0.11$      &$0.50$ &${0.51}$\\
      \hspace{-1ex}{Tiger 1}        &$0.54$   &${0.92~{\ast}}$  &${0.81~{\ast\ast}}$  &${0.35}$  &$0.48$      &$0.32$ &$0.28$\\
      \hspace{-1ex}{Tiger 2}        &${0.83~{\ast}}$  &${0.81~{\ast\ast}}$   &${0.83~{\ast}}$  &${0.14}$  &$0.51$      &$0.29$ &$0.22$\\
      \hspace{-1ex}{Coupon Book}        &${0.94~{\ast}}$   &$0.66$  &${0.69~{\ast\ast}}$   &${0.52}$ &$0.67$      &$0.57$ &$0.41$\\
      \hspace{-1ex}{Occluded Face}       & $0.79$   & $0.64$   & $0.43$   &${1.00~{\ast}}$  & $0.22$       &${0.94}$ & ${0.95~{\ast\ast}}$\\
      \hspace{-1ex}{Occluded Face 2}             & $ {0.75~{\ast\ast}}$   &$0.18$   & $ 0.60$ &${0.95~{\ast}}$ & ${0.61}$   &${0.72}$ & $0.44$   \\
      \bottomrule
      \hspace{-1ex}{\bf average precision}    &${\bf0.82~{\ast}}$ &${\bf0.69~{\ast\ast}}$ & $\bf0.64$ &${\bf0.63}$ &$\bf0.44$ &$\bf0.45$ &$\bf0.40$\\
      \bottomrule
    \end{tabular}}
  \end{minipage}
  \vspace{2ex}
\end{table*}

\subsection{Qualitative Comparison}

{\bf Heavy occlusions}.
Occlusion is one of the major issues in object tracking.
Trackers such as SCM, FragTrack and IVT are designed to resolve this problem.
Other trackers, including TLD, MIL and OAB, are less successful in handling occlusions,
especially at frames 271, 529 and 741 of the {\it Occluded Face} sequence,
and frames 176, 432 and 607 of {\it Occluded Face 2}.
SCM can obtain good performance mainly as it is capable of handling partial occlusions via a patch-based model.
The proposed AST approach can tolerate occlusions to some extent, thanks to the properties of the appearance model.
One prime example is {\it Occluded Face~2}, where AST accurately localised the severely occluded object at frame 730.

{\bf Pose Variations}.
On the {\it Tiger~2} sequence,
most trackers, including SCM, IVT and FragTrack, fail to track the object from the early frames onwards.
On {\it Tiger~2}, the proposed AST approach can accurately follow the object at frames 207 and 271 when all the other trackers have failed.
In addition, compared to the other trackers, the proposed approach partly handles motion blurring (\eg frame 344),
where the blurring is a side-effect of rapid pose variations.
On {\it Tiger~1}, although TLD obtains the best performance,
AST can successfully locate (in contrast to the other trackers)
the object at frames 204 and 249, which are subject to occlusion and severe illumination changes.

{\bf Rotations}.
The {\it Girl} and {\it Surfer} sequences include drastic out-of-plane and in-plane rotations.
On {\it Surfer}, FragTrack and SCM fail to track from the start.
The proposed AST approach consistently tracked the surfer and outperforms the other trackers.
On {\it Girl}, the IVT, OAB, and FragTrack methods fail to track in many frames.
While IVT is able to track in the beginning, it fails after frame 230.
The AST approach manages to track the correct person throughout the whole sequence,
especially towards the end where the other trackers fail due to heavy occlusion.

{\bf Illumination changes}.
The {\it Coke Can} sequence consists of dramatic illumination changes.
FragTrack fails from frame 20 where the first signs of illumination changes appear.
IVT and OAB fail from frame 40 where the frames include both severe illumination changes and slight motion blur.
MILTrack fails after frame 179 where a part of the object is almost faded by the light.
Since affine subspaces accommodate robustness to the illumination changes,
the proposed AST approach can accurately locate the object throughout the whole sequence.

{\bf Imposters/Distractors}.
The {\it Coupon Book} sequence contains a severe appearance change, as well as an imposter book to distract the tracker.
FragTrack and TLD fail mainly where the imposter book appears.
AST successfully tracks the correct book with notably better accuracy than the other methods.
~

\section{Main Findings and Future Directions}
\label{sec:conclusion}

In this paper we investigated the problem of object tracking in a video stream
where object appearance can drastically change due to factors such as occlusions and/or variations in illumination and pose.
The selection of subspaces for target representation purposes,
in addition to a regular subspace update,
are mainly driven by the need for an adaptive object template reflecting appearance changes.
We argued that modelling the appearance by affine subspaces and applying this notion
on both the object templates and the query data leads to more robustness.
Furthermore, we maintain a record of $k$ previously observed templates for a more robust tracker.

We also presented a novel subspace-to-subspace measurement approach
by reformulating the problem over Grassmann manifolds,
which provides the target representation with more robustness against intrinsic and extrinsic variations.
Finally, the tracking problem was considered as an inference task in a Markov Chain Monte Carlo framework
using particle filters to propagate sample distributions over time.

Comparative evaluation on challenging video sequences against several state-of-the-art trackers
show that the proposed AST approach obtains superior accuracy, effectiveness and consistency,
with respect to illumination changes, partial occlusions, and various appearance changes.
Unlike the other methods, AST involves no training phase.

There are several challenges, such as drifts and motion blurring, that need to be addressed.
A solution to drifts could be to formulate the update process in a semi-supervised fashion
in addition to including a training stage for the detector. 
Future research directions also include an enhancement to the updating scheme
by measuring the effectiveness of a new learned model before adding it to the bag of models.
To resolve the motion blurring issues, we can enhance the framework by introducing blur-driven models
and particle filter distributions. 
Furthermore, an interesting extension would be multi-object tracking and how to join multiple object models.

~

\section*{Acknowledgements}
\small
NICTA is funded by the Australian Government through the Department of Communications and the Australian Research Council through the ICT Centre of Excellence program.

~

\renewcommand{\baselinestretch}{1.1}\small\normalsize
\balance
{\footnotesize
\bibliographystyle{ieee}
\bibliography{references}

\begin{thebibliography}{10}\itemsep=-1pt

\bibitem{adam2006robust}
A.~Adam, E.~Rivlin, and I.~Shimshoni.
\newblock Robust fragments-based tracking using the integral histogram.
\newblock In {\em IEEE Conference on Computer Vision and Pattern Recognition
  (CVPR)}, volume~1, pages 798--805, 2006.

\bibitem{maskell2001tutorial}
M.~Arulampalam, S.~Maskell, N.~Gordon, and T.~Clapp.
\newblock A tutorial on particle filters for on-line nonlinear/non-gaussian
  bayesian tracking.
\newblock {\em IEEE Trans. Signal Processing}, 50(2):174--188, 2002.

\bibitem{babenko2009visual}
B.~Babenko, M.~Yang, and S.~Belongie.
\newblock Visual tracking with online multiple instance learning.
\newblock In {\em IEEE Conference on Computer Vision and Pattern Recognition
  (CVPR)}, pages 983--990, 2009.

\bibitem{babenko2011}
B.~Babenko, M.~Yang, and S.~Belongie.
\newblock Robust object tracking with online multiple instance learning.
\newblock {\em IEEE Transactions on Pattern Analysis and Machine Intelligence},
  33(8):1619--1632, 2011.

\bibitem{basri2011approximate}
R.~Basri, T.~Hassner, and L.~Zelnik-Manor.
\newblock Approximate nearest subspace search.
\newblock {\em IEEE Transactions on Pattern Analysis and Machine Intelligence},
  33(2):266--278, 2011.

\bibitem{birchfield1998elliptical}
S.~Birchfield.
\newblock Elliptical head tracking using intensity gradients and color
  histograms.
\newblock In {\em IEEE Conference on Computer Vision and Pattern Recognition
  (CVPR)}, pages 232--237, 1998.

\bibitem{edelman1998geometry}
A.~Edelman, T.~Arias, and S.~Smith.
\newblock The geometry of algorithms with orthogonality constraints.
\newblock {\em SIAM Journal on Matrix Analysis and Applications},
  20(2):303--353, 1998.

\bibitem{grabner2006real}
H.~Grabner, M.~Grabner, and H.~Bischof.
\newblock Real-time tracking via on-line boosting.
\newblock In {\em British Machine Vision Conference}, volume~1, pages 47--56,
  2006.

\bibitem{hamm2009extended}
J.~Hamm and D.~Lee.
\newblock Extended {G}rassmann kernels for subspace-based learning.
\newblock In {\em Advances in Neural Information Processing Systems (NIPS)},
  pages 601--608, 2009.

\bibitem{Harandi_2013_ICCV}
M.~Harandi, C.~Sanderson, C.~Shen, and B.~C. Lovell.
\newblock Dictionary learning and sparse coding on {G}rassmann manifolds: An
  extrinsic solution.
\newblock In {\em Int. Conference on Computer Vision (ICCV)}, 2013.

\bibitem{harandi2011graph}
M.~Harandi, C.~Sanderson, S.~Shirazi, and B.~C. Lovell.
\newblock Graph \mbox{embedding} discriminant analysis on {G}rassmannian
  manifolds for \mbox{improved} image set matching.
\newblock In {\em IEEE Conference on Computer Vision and Pattern Recognition
  (CVPR)}, pages 2705--2712, 2011.

\bibitem{Harandi_2013_PRL}
M.~Harandi, C.~Sanderson, S.~Shirazi, and B.~C. Lovell.
\newblock Kernel analysis on {G}rassmann manifolds for action recognition.
\newblock {\em \mbox{Pattern} \mbox{Recognition} \mbox{Letters}},
  34(15):1906--1915, 2013.

\bibitem{ho2004visual}
J.~Ho, K.~Lee, M.~Yang, and D.~Kriegman.
\newblock Visual tracking using learned linear subspaces.
\newblock In {\em IEEE Conference on Computer Vision and Pattern Recognition
  (CVPR)}, volume~1, pages 782--789, 2004.

\bibitem{isard1996contour}
M.~Isard and A.~Blake.
\newblock Contour tracking by stochastic propagation of conditional density.
\newblock {\em European Conference on Computer Vision (ECCV)}, pages 343--356,
  1996.

\bibitem{kalal2011tracking}
Z.~Kalal, K.~Mikolajczyk, and J.~Matas.
\newblock Tracking-learning-detection.
\newblock {\em IEEE Transactions on Pattern Analysis and Machine Intelligence},
  34(7):1409--1422, 2012.

\bibitem{kim2007discriminative}
T.~Kim, J.~Kittler, and R.~Cipolla.
\newblock Discriminative learning and recognition of image set classes using
  canonical correlations.
\newblock {\em IEEE Transactions on Pattern Analysis and Machine Intelligence},
  29(6):1005--1018, 2007.

\bibitem{kittler1998combining}
J.~Kittler, M.~Hatef, R.~Duin, and J.~Matas.
\newblock On combining classifiers.
\newblock {\em IEEE Transactions on Pattern Analysis and Machine Intelligence},
  20(3):226--239, 1998.

\bibitem{li2012incremental}
X.~Li, A.~Dick, C.~Shen, A.~van~den Hengel, and H.~Wang.
\newblock Incremental learning of {3D-DCT} compact representations for robust
  visual tracking.
\newblock {\em IEEE Transactions on Pattern Analysis and Machine Intelligence},
  35(4):863--881, 2013.

\bibitem{ross2008incremental}
D.~Ross, J.~Lim, R.~Lin, and M.~Yang.
\newblock Incremental learning for robust visual tracking.
\newblock {\em Int. Journal of Computer Vision (IJCV)}, 77(1):125--141, 2008.

\bibitem{Sanderson_AVSS_2012}
C.~Sanderson, M.~Harandi, Y.~Wong, and B.~C. Lovell.
\newblock Combined learning of salient local descriptors and distance metrics
  for image set face verification.
\newblock In {\em IEEE International Conference on Advanced Video and
  Signal-Based Surveillance (AVSS)}, pages 294--299, 2012.

\bibitem{Shirazi_2012_ICIP}
S.~Shirazi, M.~Harandi, C.~Sanderson, A.~Alavi, and B.~C. Lovell.
\newblock Clustering on {G}rassmann manifolds via kernel embedding with
  \mbox{application} to action analysis.
\newblock In {\em Int. Conference on Image \mbox{Processing} (ICIP)}, pages
  781--784, 2012.

\bibitem{turaga2011statistical}
P.~Turaga, A.~Veeraraghavan, A.~Srivastava, and R.~Chellappa.
\newblock Statistical computations on {G}rassmann and {S}tiefel manifolds for
  image and video-based recognition.
\newblock {\em IEEE Transactions on Pattern Analysis and Machine Intelligence},
  33(11):2273--2286, 2011.

\bibitem{von2007tutorial}
U.~Von~Luxburg.
\newblock A tutorial on spectral clustering.
\newblock {\em Statistics and Computing}, 17(4):395--416, 2007.

\bibitem{wang2011superpixel}
S.~Wang, H.~Lu, F.~Yang, and M.-H. Yang.
\newblock Superpixel tracking.
\newblock In {\em Int. Conference on Computer Vision (ICCV)}, pages 1323--1330,
  2011.

\bibitem{wang2008online}
T.~Wang, A.~Backhouse, and I.~Gu.
\newblock Online subspace learning on {G}rassmann manifold for moving object
  tracking in video.
\newblock In {\em IEEE International Conference on Acoustics, Speech and Signal
  Processing (ICASSP)}, pages 969--972, 2008.

\bibitem{zhong2012}
W.~Zhong, H.~Lu, and M.-H. Yang.
\newblock Robust object tracking via sparsity-based collaborative model.
\newblock In {\em IEEE Conference on Computer Vision and Pattern Recognition
  (CVPR)}, pages 1838--1845, 2012.

\end{thebibliography}
}

\end{document}